\documentclass[graybox]{svmult}

\usepackage{type1cm}        
\usepackage{makeidx}         
\usepackage{graphicx}        
\usepackage{epstopdf}        
\usepackage{multicol}        
\usepackage[bottom]{footmisc}

\usepackage{newtxtext}       %
\usepackage[varvw]{newtxmath}       
\usepackage{algorithmic}
\usepackage{graphicx}
\usepackage{balance}
\usepackage{textcomp}
\usepackage{algorithm}
\usepackage{algorithmic}
\usepackage{newtxmath}
\usepackage{subcaption}
\usepackage{booktabs}
\usepackage{multirow}
\usepackage{amsmath}
\usepackage{adjustbox}

\newcommand\algorithmicprocedure{\textbf{procedure}}
\newcommand{\algorithmicendprocedure}{\algorithmicend\ \algorithmicprocedure}
\makeatletter
\newcommand\PROCEDURE[3][default]{%
  \ALC@it
  \algorithmicprocedure\ \textsc{#2}(#3)%
  \ALC@com{#1}%
  \begin{ALC@prc}%
}
\newcommand\ENDPROCEDURE{%
  \end{ALC@prc}%
  \ifthenelse{\boolean{ALC@noend}}{}{%
    \ALC@it\algorithmicendprocedure
  }%
}
\newenvironment{ALC@prc}{\begin{ALC@g}}{\end{ALC@g}}

\begin{document}
\title{Detection of T-shirt Presentation Attacks in Face Recognition Systems}
\title*{Detection of T-shirt Presentation Attacks in Face Recognition Systems}


\author{Mathias Ibsen, Loris Tim Ide,  Christian Rathgeb, and Christoph Busch}


\institute{Mathias Ibsen \at Computer Science, Hochschule Darmstadt, Darmstadt, Germany \\ \email{mathias.ibsen@h-da.de}
\and Loris Tim Ide \at Computer Science, Hochschule Darmstadt, Darmstadt, Germany \\ \email{loristim.ide@stud.h-da.de}
\and Christian Rathgeb \at Computer Science, Hochschule Darmstadt, Darmstadt, Germany \\ \email{christian.rathgeb@h-da.de}
\and Christian Rathgeb \at Computer Science, Hochschule Darmstadt, Darmstadt, Germany \\ \email{christian.rathgeb@h-da.de}
\and Christoph Busch \at Computer Science, Hochschule Darmstadt, Darmstadt, Germany \\ \email{christoph.busch@h-da.de}}

%

%
\maketitle


\abstract*{Face recognition systems are often used for biometric authentication. Nevertheless, it is known that without any protective measures, face recognition systems are vulnerable to presentation attacks. To tackle this security problem, methods for detecting presentation attacks have been developed and shown good detection performance on several benchmark datasets. However, generalising presentation attack detection methods to new and novel types of attacks is an ongoing challenge. In this work, we employ 1,608 T-shirt attacks of the T-shirt Face Presentation Attack (TFPA) database using 100 unique presentation attack instruments together with 152 bona fide presentations. In a comprehensive evaluation, we show that this type of attack can compromise the security of face recognition systems. Furthermore, we propose a detection method based on spatial consistency checks in order to detect said T-shirt attacks. Precisely, state-of-the-art face and person detectors are combined to analyse the spatial positions of detected faces and persons based on which T-shirt attacks can be reliably detected.}

\abstract{Face recognition systems are often used for biometric authentication. Nevertheless, it is known that without any protective measures, face recognition systems are vulnerable to presentation attacks. To tackle this security problem, methods for detecting presentation attacks have been developed and shown good detection performance on several benchmark datasets. However, generalising presentation attack detection methods to new and novel types of attacks is an ongoing challenge. In this work, we employ 1,608 T-shirt attacks of the T-shirt Face Presentation Attack (TFPA) database using 100 unique presentation attack instruments together with 152 bona fide presentations. In a comprehensive evaluation, we show that this type of attack can compromise the security of face recognition systems. Furthermore, we propose a detection method based on spatial consistency checks in order to detect said T-shirt attacks. Precisely, state-of-the-art face and person detectors are combined to analyse the spatial positions of detected faces and persons based on which T-shirt attacks can be reliably detected.}

\section{Introduction}

Biometric systems are systems that recognise individuals on the basis of observing biological and behavioural characteristics. One particular type of biometric system is face recognition\index{face recognition}, which uses the characteristics of faces to recognise people. Face recognition systems are easy to use as faces can be captured remotely using commercially available hardware. Furthermore, state-of-the face recognition systems have achieved near perfect performance on many challenging benchmarks~\cite{Deng-ArcFace-IEEE-CVPR-2019, Meng-MagFace-CVPR-2021}. As a result, such face recognition systems are used in a wide range of applications, including personal, industrial and governmental applications. Although biometric recognition performance of face recognition systems has significantly increased in recent years~\cite{Zhao-FaceRecognSurvey-2003,Abate-2DAnd3DFaceRecognition-2007,LiJain-HandbookOfFaceRecognition-2011}, it has been shown that these systems are vulnerable to presentation attacks (PAs)\index{presentation attacks (PAs)}, e.g.\ in~\cite{Chingovska-OnTheEffectivenessOfLocalBinaryPatternInFaceAntgiSpoofing-BIOSIG-2012,Raghavendra-FacePAD-Survey-2017}. In ISO/IEC 30107-3\index{ISO/IEC 30107-3}~\cite{ISO-IEC-30107-3-PAD-metrics-2023} a PA is defined as a "\textit{presentation to the biometric data capture subsystem with the goal of interfering with the operation of the biometric system}".

To decrease the vulnerability of face recognition systems with respect to PAs, hardware- and software-based detection methods have been proposed along with several benchmark datasets. Early works were mostly focused on detecting print and replay attacks and was often trained and evaluated on a single or a few related types of PAs. Later works have introduced more sophisticated attacks, e.g.\ silicon masks~\cite{Raghavendra-CustomSiliconFaceMask-VulnerabilityAndPAD-IWBF-2019} and makeup attacks~\cite{Drozdowski-MakeupPADataset-IWBF-2021} and the focus of many recent works proposing presentation attack detection (PAD)\index{presentation attack detection (PAD)} systems has been on generalisation to unknown attacks. In such a more realistic detection scenario where the attack has not been seen during training, researchers have reported significantly higher error rates compared to evaluating PAD for known or similar attack types~\cite{GonzalezSoler-UnkownAttacksFace-IET-Biometrics-2021} which illustrate that creating robust PAD systems for real-world applications remains an open challenge. \par PAD systems were initially based on hand-crafted features and simple machine learning techniques, but have since evolved into deep learning-based methods. Recent advances include domain adaptation~\cite{Wang-PAD-AdvDomainAdaptation-TIFS-2020}, anomaly detection~\cite{Nikisins-OnEffectivenessOfAnamolyDetectionApproachesAgainstUnseenPresentationAttacksInFaceAntiSpoofing-ICB-2018, Fatemifar-SpoofingAttackDetectionByAnomalyDetection-2019, Ibsen-DifferentialAnomalyDetectionForFacialImages-WIFS-2021}, and new loss functions to better modulate the contribution of individual channels in multi-stream architectures~\cite{George-CMFLForRGBDFaceAntiSpoofing-CVPR-2021}. Despite some progress, generalisability to unknown PAs remains an open problem~\cite{Nikisins-OnEffectivenessOfAnamolyDetectionApproachesAgainstUnseenPresentationAttacksInFaceAntiSpoofing-ICB-2018, CostaPazo-FacePADComrehensiveEvaluationGeneralisationProblem-IET-2021}, especially for methods that do not rely on specific hardware-based approaches. 

\begin{figure}[!t]
\centering
\includegraphics[width=\linewidth]{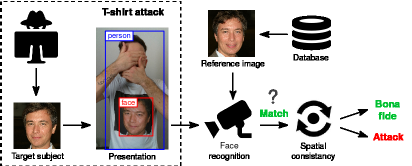}
\caption{Can T-shirts with faces printed on them be used to attack face recognition systems? The figure illustrates the attack scenario, which relates, for instance, to a verification attempt in an automated border control gate. Adapted from~\cite{Ibsen-AttackingFaceRecognitionWithTshirts-IEEEAccess-2023}}
\label{fig:tshirt_pa_intro}
\end{figure}

In the context of this work, T-shirt face presentation attacks\index{T-shirt face presentation attacks}, i.e.\ T-shirts with a human face printed on them, are of particular interest as illustrated in Figure~\ref{fig:tshirt_pa_intro}. These T-shirt attacks have already been identified by border authorities as a potential instrument for PAs~\cite{EU-Frontex-TechnicalGuideEESRelatedEquipment-2021, BSI-BiometricsPublicSectorApplicationsPart3-2023}. The fabrication of the presentation attack instrument\index{presentation attack instrument} is cheap compared to other types of presentation attack instruments, e.g.\ 3D silicone masks. Furthermore, PAs with a facial image on clothing have two major advantages: concealed use is easier because the facial image does not have to be held with the hands towards the camera. In addition, the T-shirt can be worn under a jacket, which can only be opened immediately before the verification attempt. If the jacket is only opened slightly and held open at the front, the facial image can only be seen from the front. This means that the risk when used in a supervised authentication system is significantly lower than, for instance, when using a facial image on paper or a display.
Furthermore, the facial image on an item of clothing can easily be given a 3D shape, e.g.\ by placing an object or hand underneath, which can fool 3D-based PAD systems.

The feasibility of T-shirt concealment attacks, in which an attacker attempts to conceal his own identity and avoid being recognised, was demonstrated in~\cite{Xu-AdversarialTshirtAttack-2020}. In this work, adversarial patterns were printed on T-shirts to avoid automatic detection of persons. However, in contrast to this previous work, our work focuses on \textit{impersonation attacks}\index{impersonation attacks}, i.e.\ where an attacker tries to be authenticated as another target identity. To carry out this attack, an attacker prints a facial image of a target identity on a T-shirt and, in some verification scenarios, the attacker would also need to get hold of an ID-document of the target identity. The T-shirt is then worn normally and presented by the attacker in front of the biometric capture device, ideally obscuring their real face, e.g.\ with a face mask or their hands. For the attack to be successful, the face recognition system must detect and process the face on the T-shirt and compare the extracted features with those of the target identity, which gives a similarity score that exceeds the system's decision threshold. This attack scenario is illustrated in Figure~\ref{fig:tshirt_pa_intro} and has been investigated in~\cite{Ibsen-AttackingFaceRecognitionWithTshirts-IEEEAccess-2023}. This chapter builds upon the work in~\cite{Ibsen-AttackingFaceRecognitionWithTshirts-IEEEAccess-2023} utilizing the T-shirt Face Presentation Attack (TFPA) database\index{T-shirt Face Presentation Attack (TFPA) database}, including T-shirt PAs based on synthetically generated facial images, each worn by at least two people in eight different recording scenarios. The T-shirt PAs are available for researchers\footnote{\url{https://dasec.h-da.de/research/biometrics/t-shirt-face-presentation-attack-tfpa-database}}. In addition, bona fide presentations\index{bona fide presentations} of 19 individuals are captured in the same acquisition scenario. The feasibility of launching the T-shirt PAs is demonstrated in experiments. Furthermore, a PAD algorithm is introduced which analyses the spatial relation between faces and persons detected in a face recognition system. On the aforementioned database, it is showcased that the presented PAD algorithm is able to reliably detect T-shirt PAs.

\section{Related Works}

This section presents an overview of face PAs based on clothing and accessories and vulnerability studies on face recognition systems against PAs.

\subsection{Presentation Attacks through Clothing and Accessories}

Clothing and accessories can be used as attack instrument against face recognition systems. Most known attacks for clothing or accessories focus on imprinting adversarial patterns on physical objects that can be worn or attached to the face. In~\cite{Sharif-FrameworkAdversarialExamples-ACM-2019}, the authors illustrate the use of generative networks for obfuscation and impersonation attacks on face recognition systems by imprinting adversarial patterns on physical glasses. The experiments only considered a small test set but the obtained results showed that it was possible to make successful obfuscation attacks where, in the worst case, 81\% of the images in a recorded video were classified as another identity. For impersonation, worse results were achieved where an average of 67.58\% of video frames were classified as the target identity, indicating that in some cases, the proposed glasses can be used for successfully impersonating another identity. In~\cite{Xu-AdversarialTshirtAttack-2020}, the authors proposed to apply adversarial patterns to T-shirts. The aim of the attack patterns was to obfuscate, or more precisely, to avoid detection by a modern person detection algorithm. The results showed an attack success rate of 57\% in the physical domain using the YOLOv2 algorithm. In~\cite{Ryu-AdversarialAttacksByAttachingNoiseMarkersAgainstFR-2021-JISA}, the authors showed that by knowing which face recognition system is used, it is possible to add noise markers to a face such that it in some cases can be used to successfully perform impersonation attacks.

Further works have demonstrated that makeup can hamper face recognition accuracy~\cite{Dantcheva-CanFacialCosmeticsAffectTheMatchingAccuracyOfFaceRecognitionSystems} and even be used for impersonation attacks~\cite{Chen-SpoofingFacesUsingMakeupAnInvestigativeStudy-IEEE-2017}. In~\cite{Rathgeb-MakeupAttackDetection-IWBF-2020, Drozdowski-MakeupPADataset-IWBF-2021} it was found that, especially, high quality makeup-induced impersonation attacks can be used to successfully attack face recognition systems.

More recently, the attack potential of T-shirt PAs on face recognition system has been investigated in~\cite{Ibsen-AttackingFaceRecognitionWithTshirts-IEEEAccess-2023}. It was shown that state-of-the-art face recognition systems are highly vulnerable to this type of PA. Moreover, generic PAD methods analysing captured attacks in various spectra were found to reveal only moderate detection performance.  

\subsection{Vulnerability of Face Recognition to Presentation Attacks}

Several papers have investigated the effects of different PAs on face recognition performance. In~\cite{Chingovska-BiometricEvaluationUnderSpoofingAttacks-TIFS-2014,Raghavendra-FacePAD-Survey-2017,Mohammadi-DeeplyVulnerableAStudy-IET-2017}, various researchers studied the vulnerability of face recognition systems against print and replay attacks. These works differ in the way the security thresholds of the systems were selected and used to evaluate the vulnerability of face recognition systems. However, they all demonstrate the feasibility of print and replay attacks. Strictly speaking, success rates in terms of impostor attack presentation accept rate (IAPAR)~\cite{ISO-IEC-30107-3-PAD-metrics-2023} of over 90\% have been reported, with a few exceptions, notably the inter-session variability algorithm in~\cite{Mohammadi-DeeplyVulnerableAStudy-IET-2017}. However, as the authors note, this system has a high false non-match rate (FNMR) on bona fide images with the same threshold, indicating its limited recognition performance in general. The results suggest that face recognition systems with higher performance on bona fide images are more prone to PAs. In~\cite{Erdogmus-Spoofing2DFRwith3DMasks-BIOSIG-2013,Raghavendra-PADApplicationTo3DFaceMask-ICIP-2014,Erdogmus_3DMAD_BTAS_2013} the authors show the vulnerability of face recognition systems to 3D masks, which have a high vulnerability at the selected thresholds, i.e.\ $>$78\% IAPAR, for most algorithms except for the inter-session variability algorithm evaluated in~\cite{Erdogmus_3DMAD_BTAS_2013}, where an IAPAR of 65.7\% is achieved. In a more recent work~\cite{Raghavendra-CustomSiliconFaceMask-VulnerabilityAndPAD-IWBF-2019}, using a new database of custom silicone masks for eight subjects, less significant vulnerability was found for 3D masks using two commercial systems. In the worst case, which corresponds to a false acceptance rate (FAR) of 0.1\%, an IAPAR of 28.20\% was determined. At a lower FAR of 0.01\%, the face recognition systems were shown to prevent 3D mask PAs.

\begin{figure}[!tb]
\begin{subfigure}{0.24\linewidth}
    \centering 
  \includegraphics[width=\textwidth]{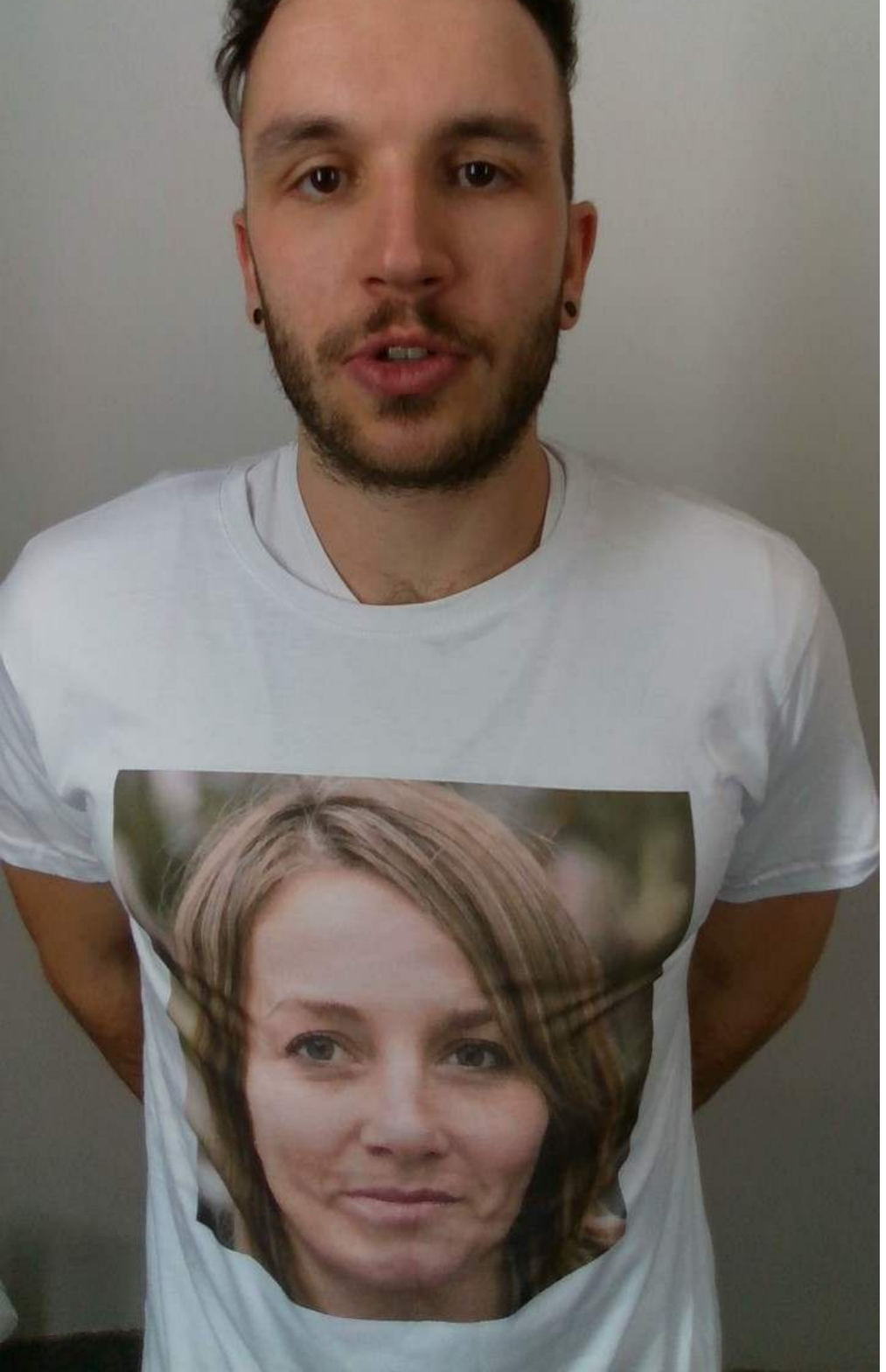}
  \caption{normal}
\end{subfigure}
\begin{subfigure}{0.24\linewidth}
    \centering 
  \includegraphics[width=\textwidth]{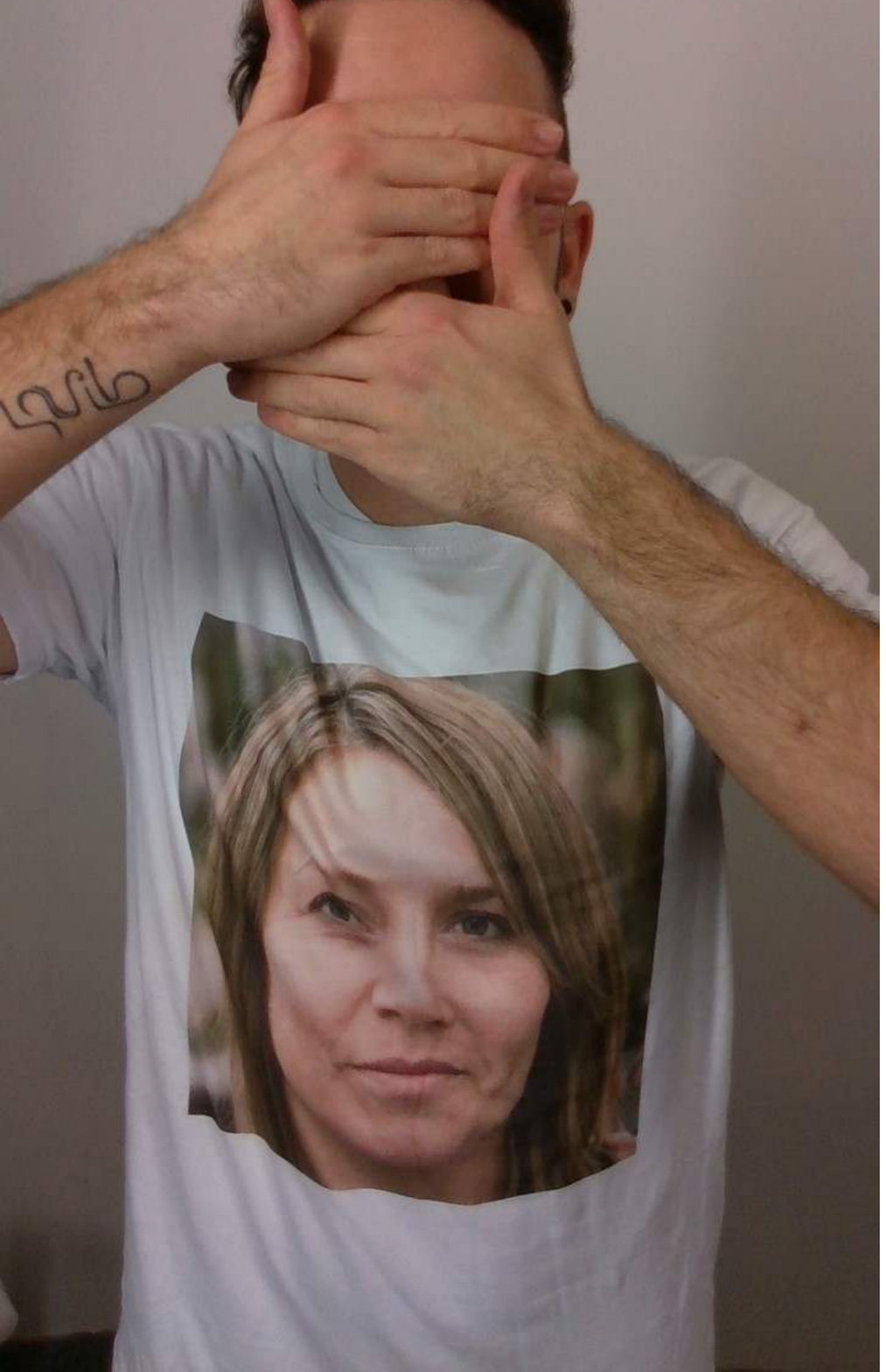}
  \caption{covered}
\end{subfigure}
\begin{subfigure}{0.24\linewidth}
    \centering 
  \includegraphics[width=\textwidth]{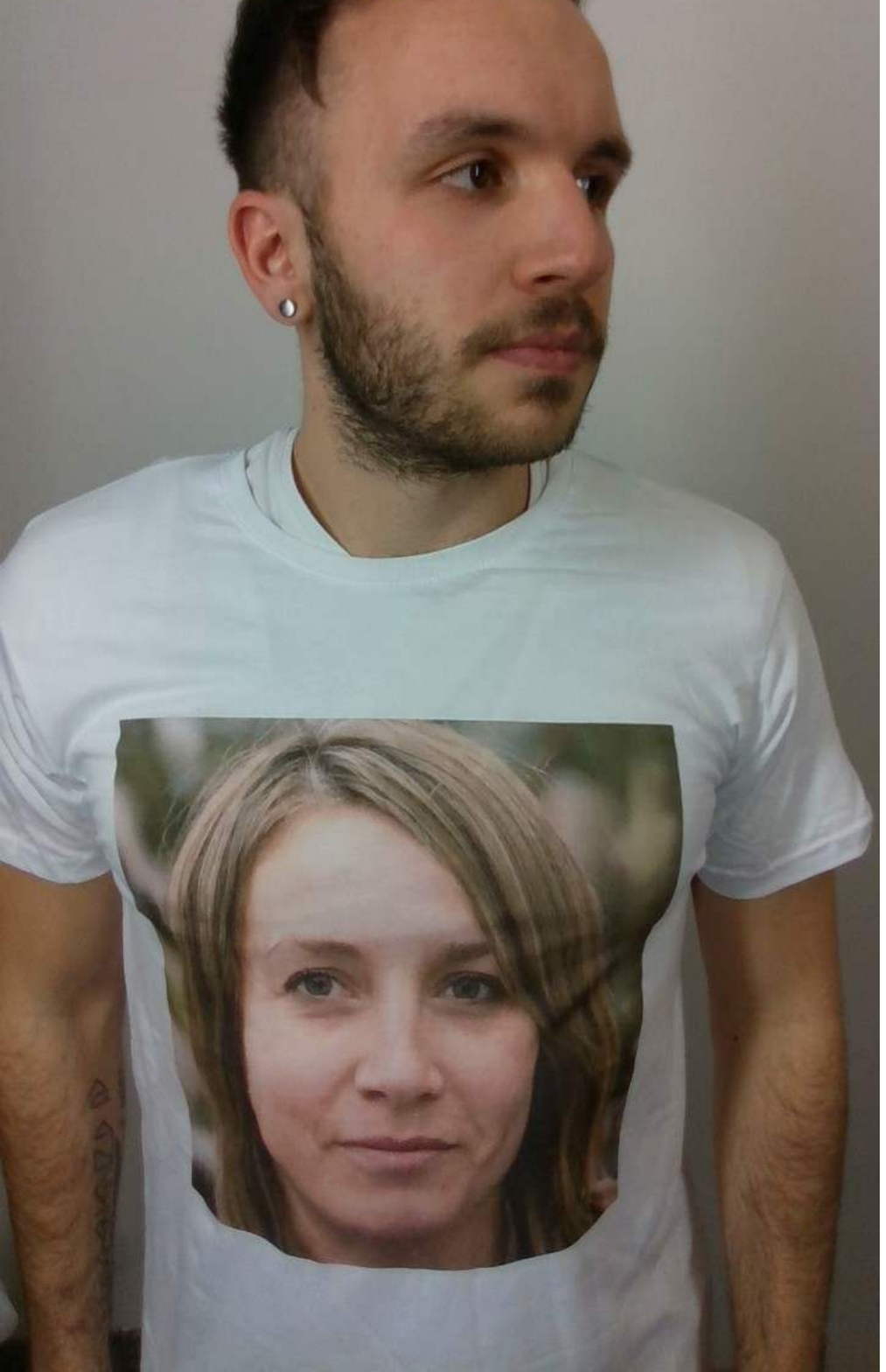}
  \caption{left}
\end{subfigure}
\begin{subfigure}{0.24\linewidth}
    \centering 
  \includegraphics[width=\textwidth]{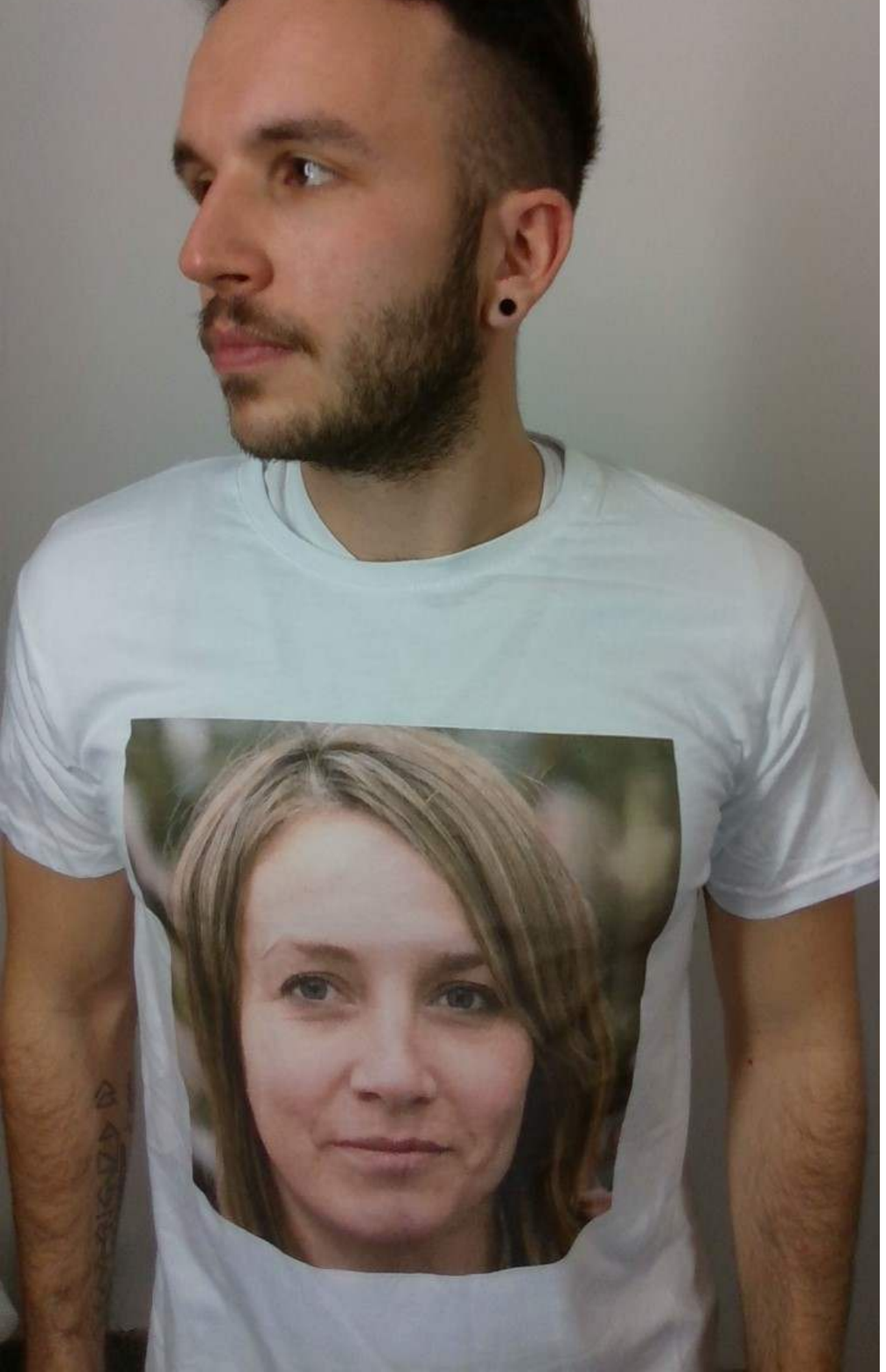}
  \caption{right}
\end{subfigure}\\
\begin{subfigure}{0.24\linewidth}
    \centering 
  \includegraphics[width=\textwidth]{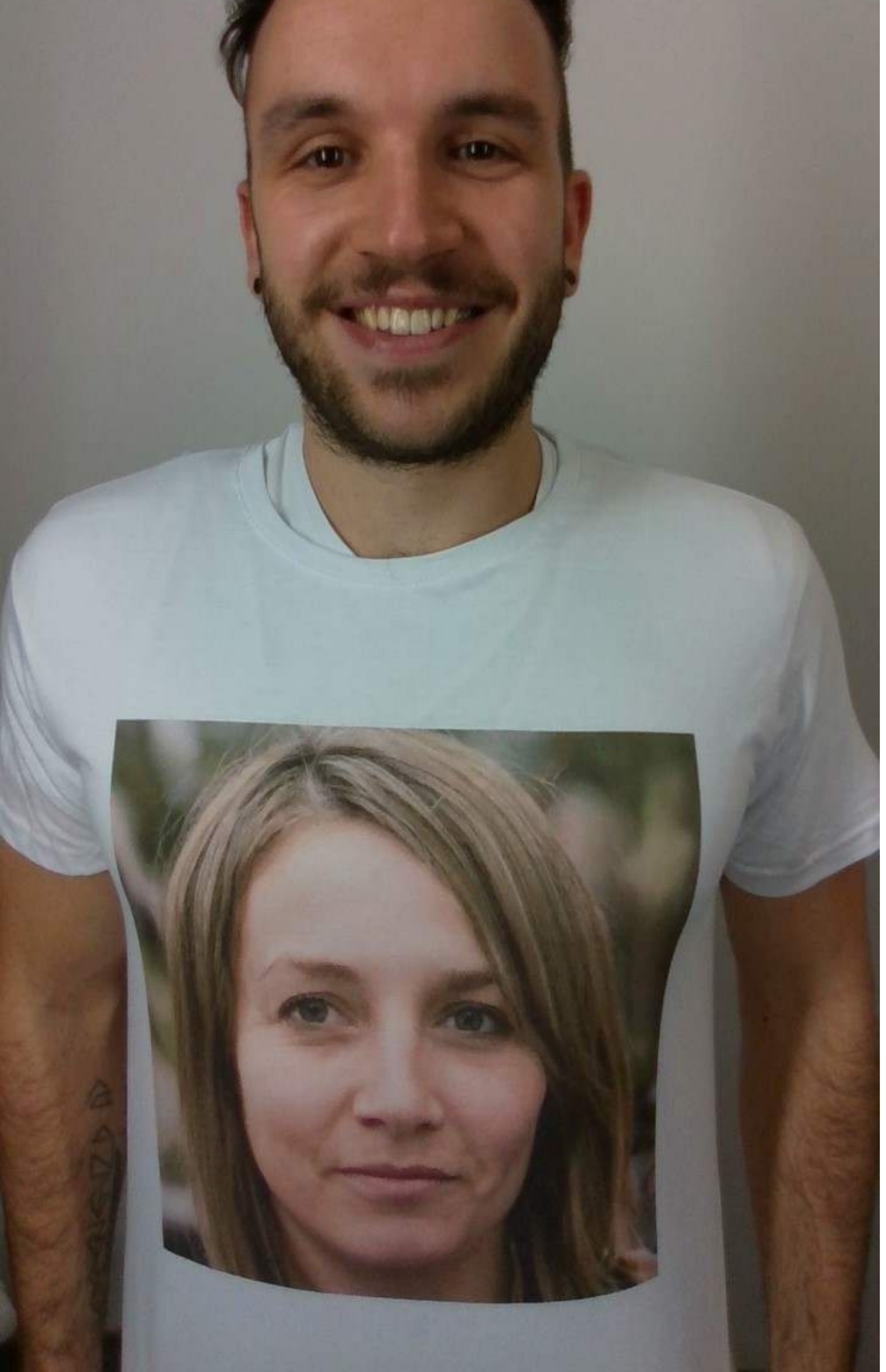}
  \caption{strech}
\end{subfigure}
\begin{subfigure}{0.24\linewidth}
    \centering 
  \includegraphics[width=\textwidth]{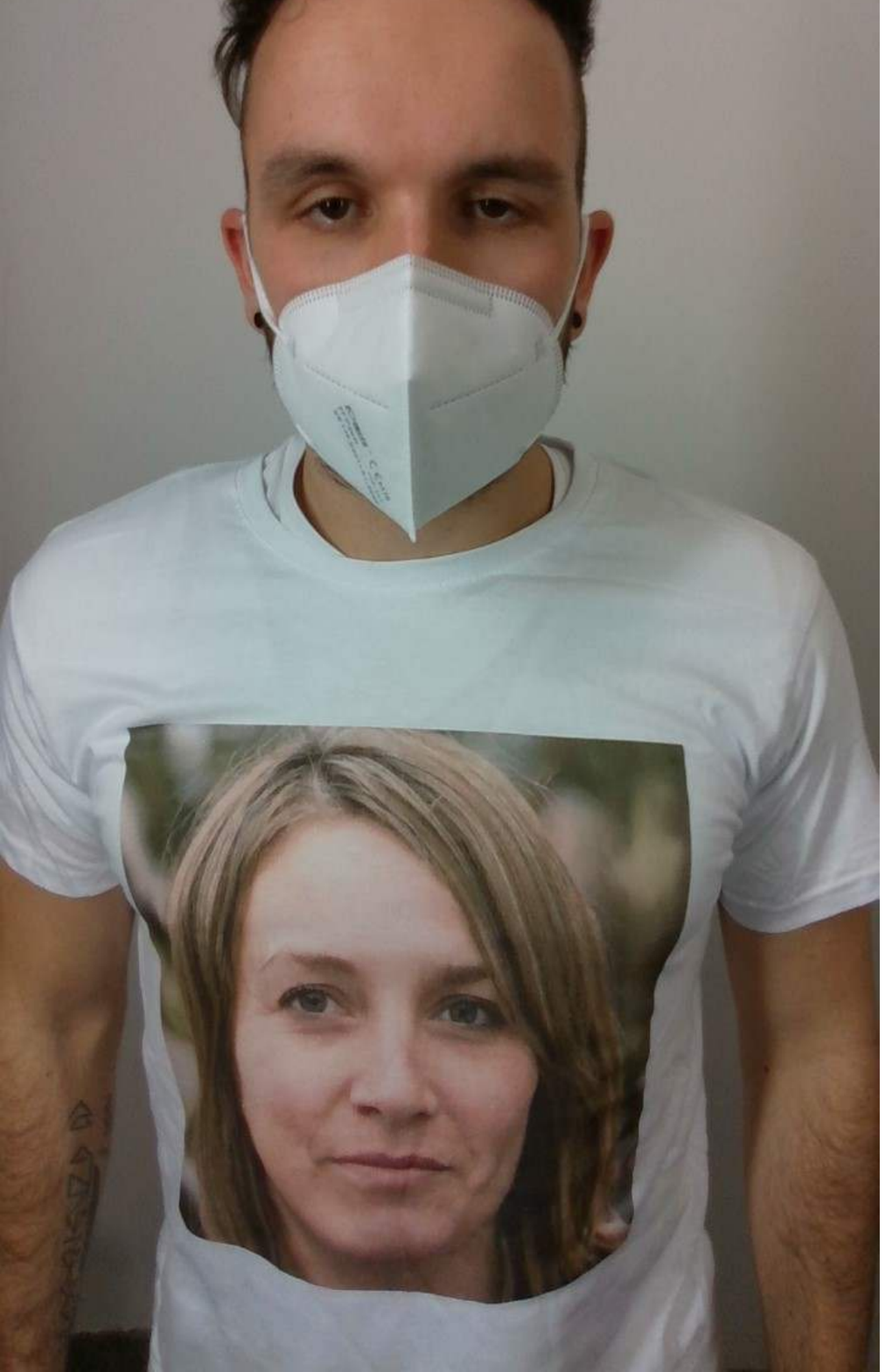}
  \caption{mask}
\end{subfigure}
\begin{subfigure}{0.24\linewidth}
    \centering 
  \includegraphics[width=\textwidth]{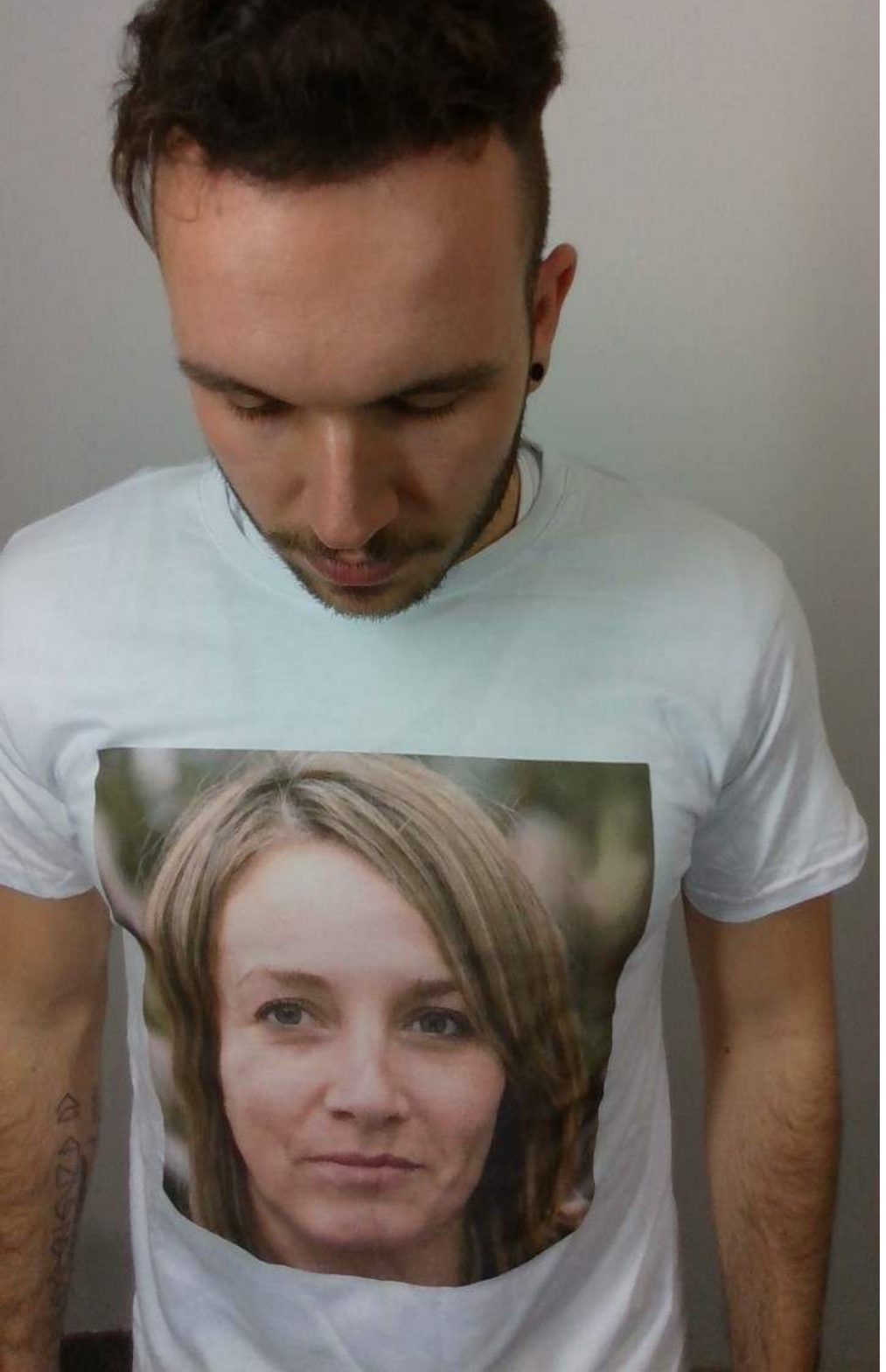}
  \caption{down}
\end{subfigure}
\begin{subfigure}{0.24\linewidth}
    \centering 
  \includegraphics[width=\textwidth]{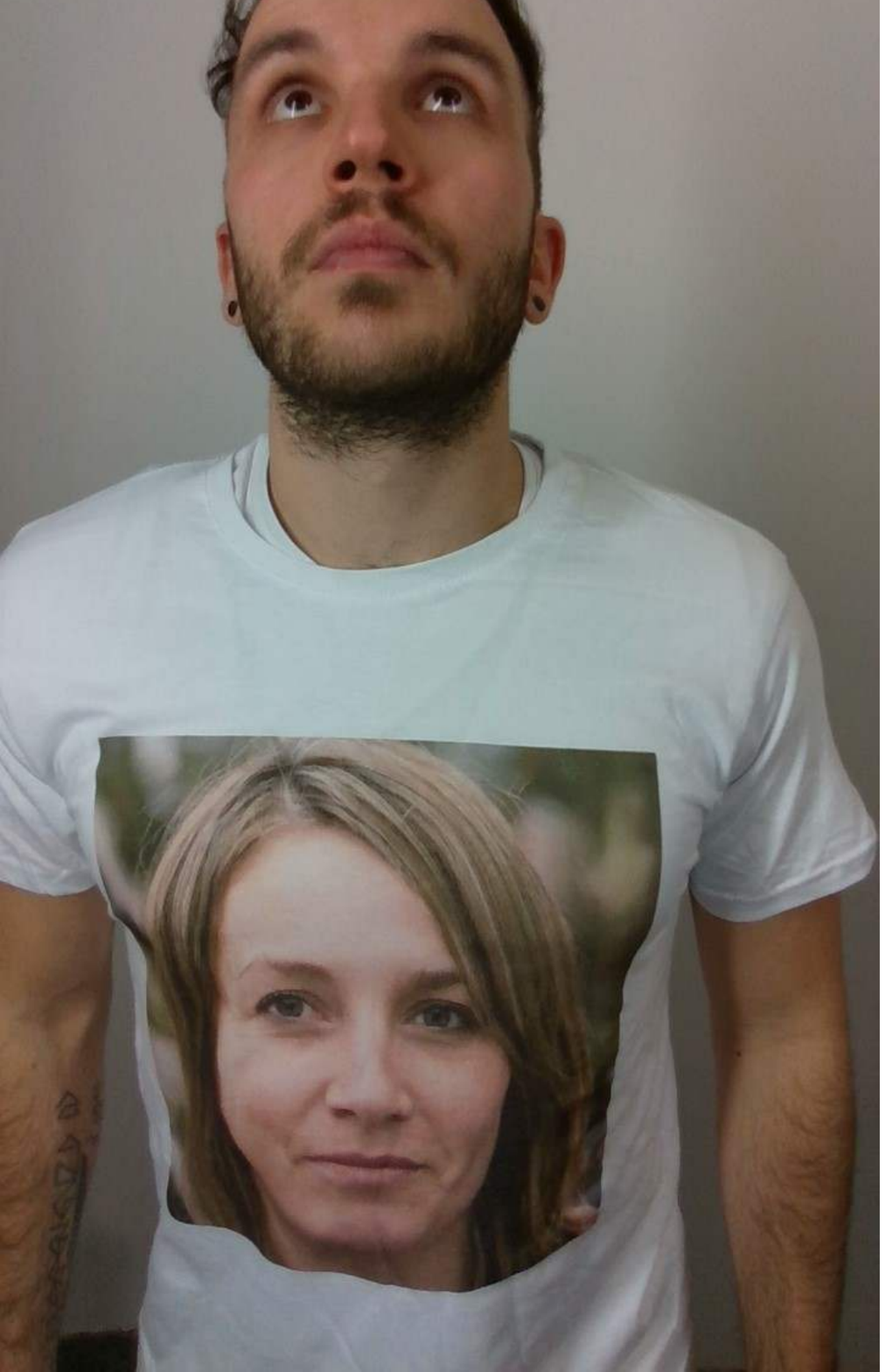}
  \caption{up}
\end{subfigure}
\caption{Example images of different T-shirt PAs of the TFPA database for a single data subject}
\label{fig:capturing_scenarios}
\end{figure}

\section{T-shirt Face Presentation Attack Database}

We use the T-shirt Face Presentation Attack (TFPA) database which consists of 1,608 T-shirt PAs taken with 100 unique T-shirts with facial images printed on them. A total of eight subjects participated in the data collection, with each T-shirt being worn by at least two different individuals. For each T-shirt worn by a subject, the subject was asked to present the T-shirt across different poses, resulting in eight different recording scenarios per T-shirt worn by a subject. Both visible (RGB) and depth data were collected. However, depth data is not used in this work. Examples of T-shirt PAs taken from TFPA can be seen in Figure~\ref{fig:capturing_scenarios}.  In addition, bona fide presentations of 19 subjects have been captured in the same poses and capturing scenarios. That is, the Intel RealSense Depth Camera D435 was used for data acquisition. The Intel RealSense Depth Camera D435 is an RGB-D sensor consisting of a pair of depth sensors, an RGB sensor and an infrared projector. It can capture depth images with an action radius of up to 10 metres and a depth image rate of up to 90 images per second. It is customisable with the Intel RealSense SDK 2.0. All captured RGB and depth images have a resolution of 1280 x 720 pixels.

\begin{figure}[!tb]
\begin{subfigure}{0.24\linewidth}
    \centering 
  \includegraphics[width=\textwidth]{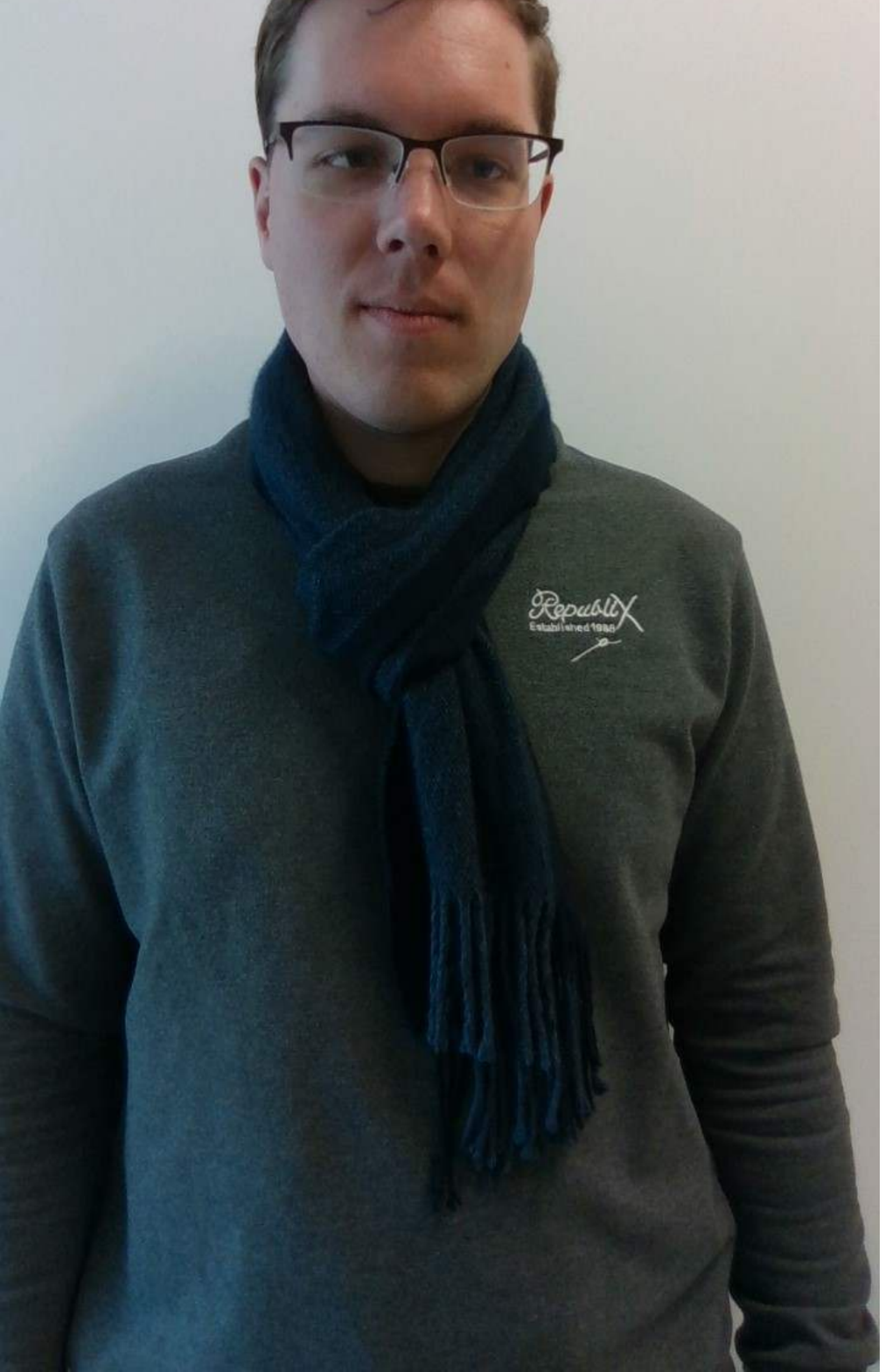}
  \caption{normal}
\end{subfigure}
\begin{subfigure}{0.24\linewidth}
    \centering 
  \includegraphics[width=\textwidth]{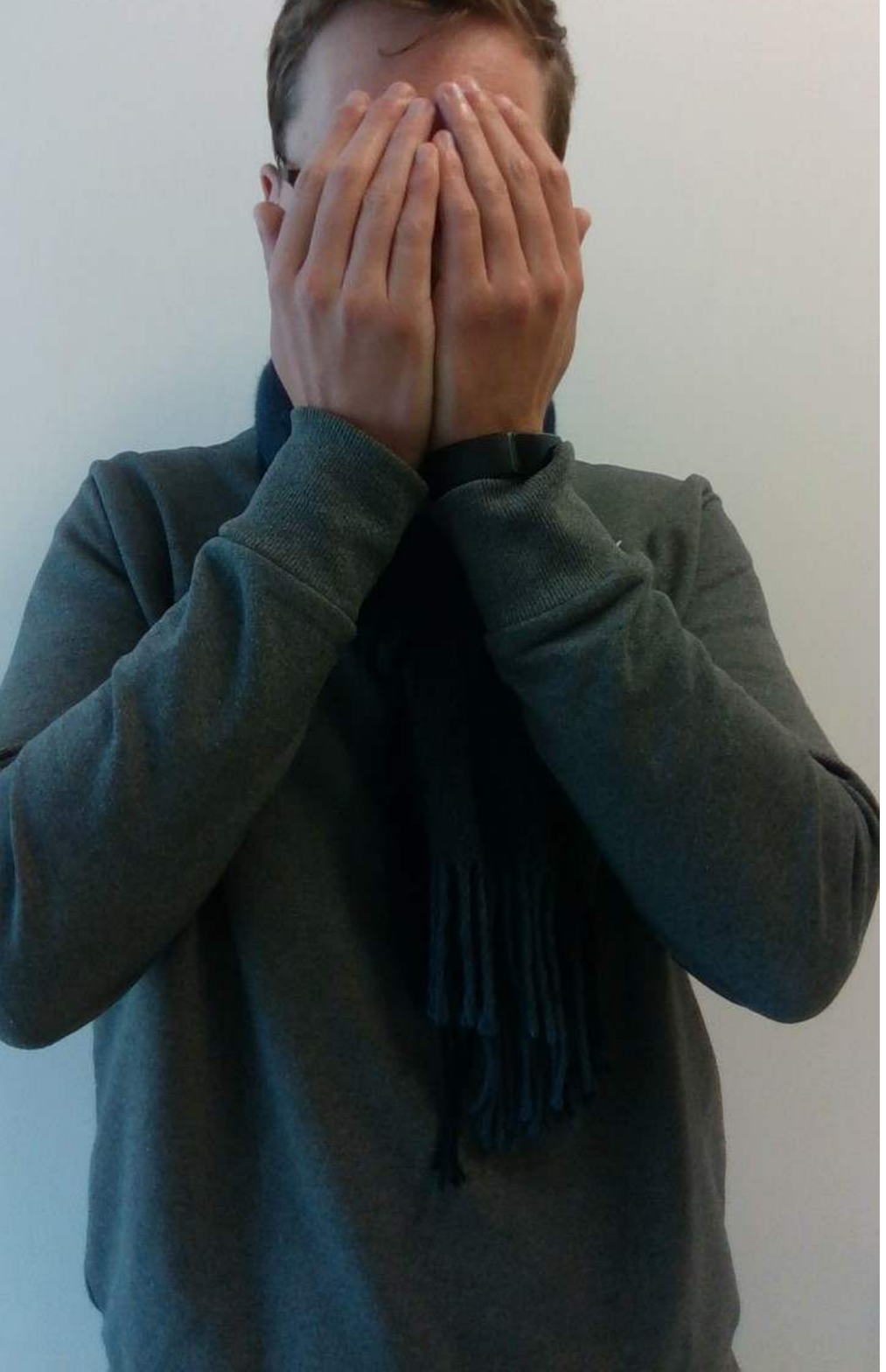}
  \caption{covered}
\end{subfigure}
\begin{subfigure}{0.24\linewidth}
    \centering 
  \includegraphics[width=\textwidth]{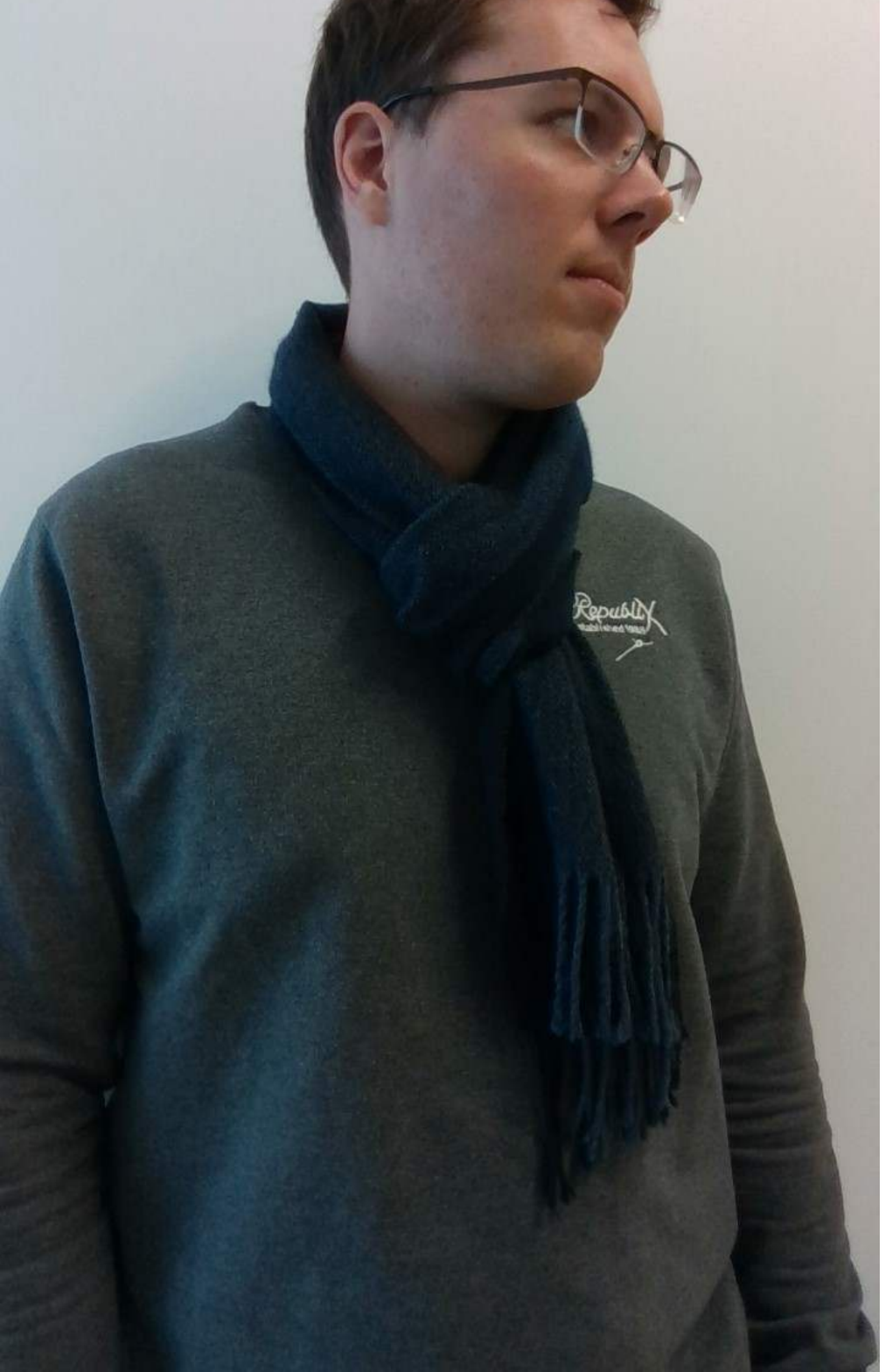}
  \caption{left}
\end{subfigure}
\begin{subfigure}{0.24\linewidth}
    \centering 
  \includegraphics[width=\textwidth]{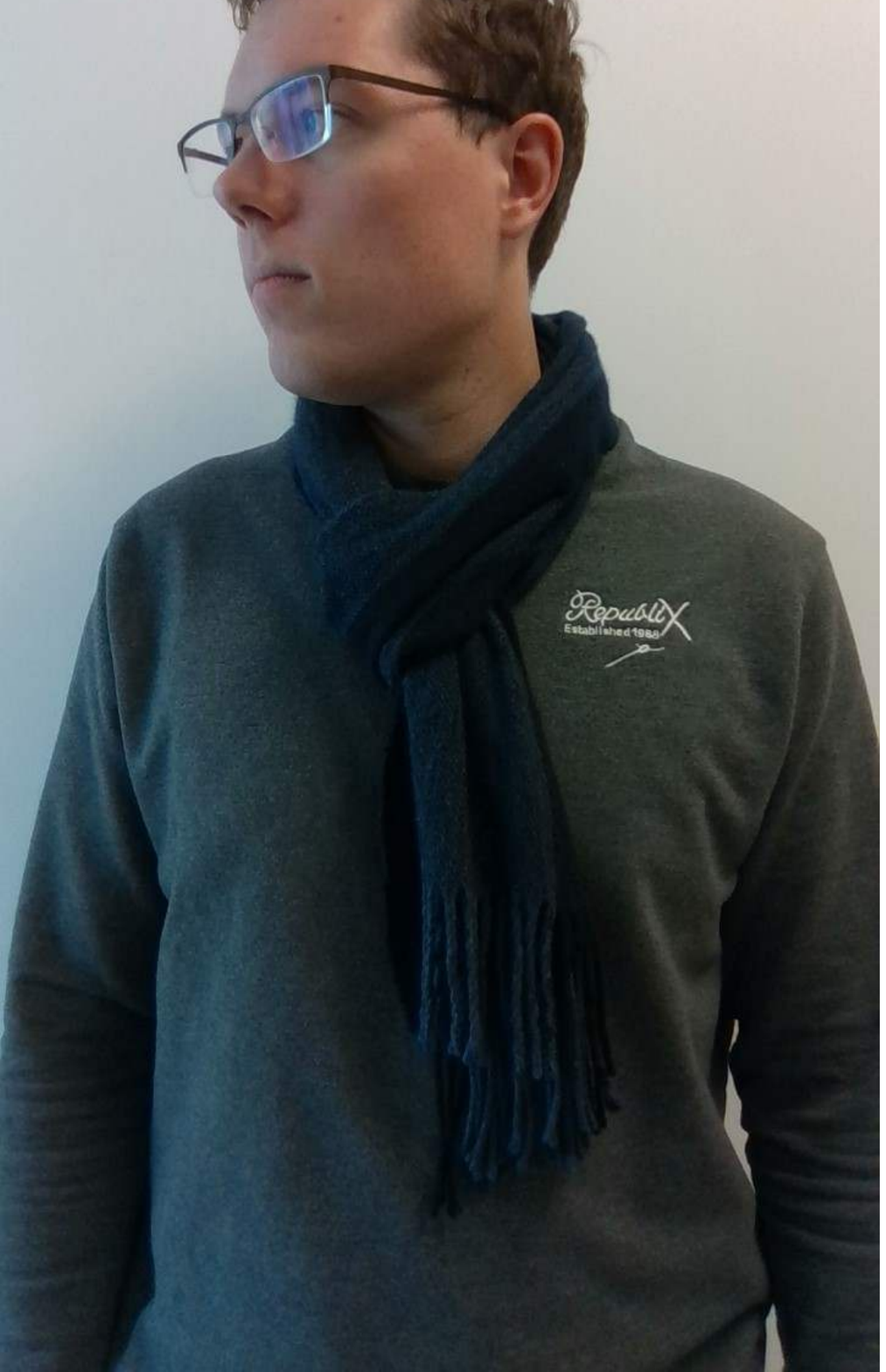}
  \caption{right}
\end{subfigure}\\
\begin{subfigure}{0.24\linewidth}
    \centering 
  \includegraphics[width=\textwidth]{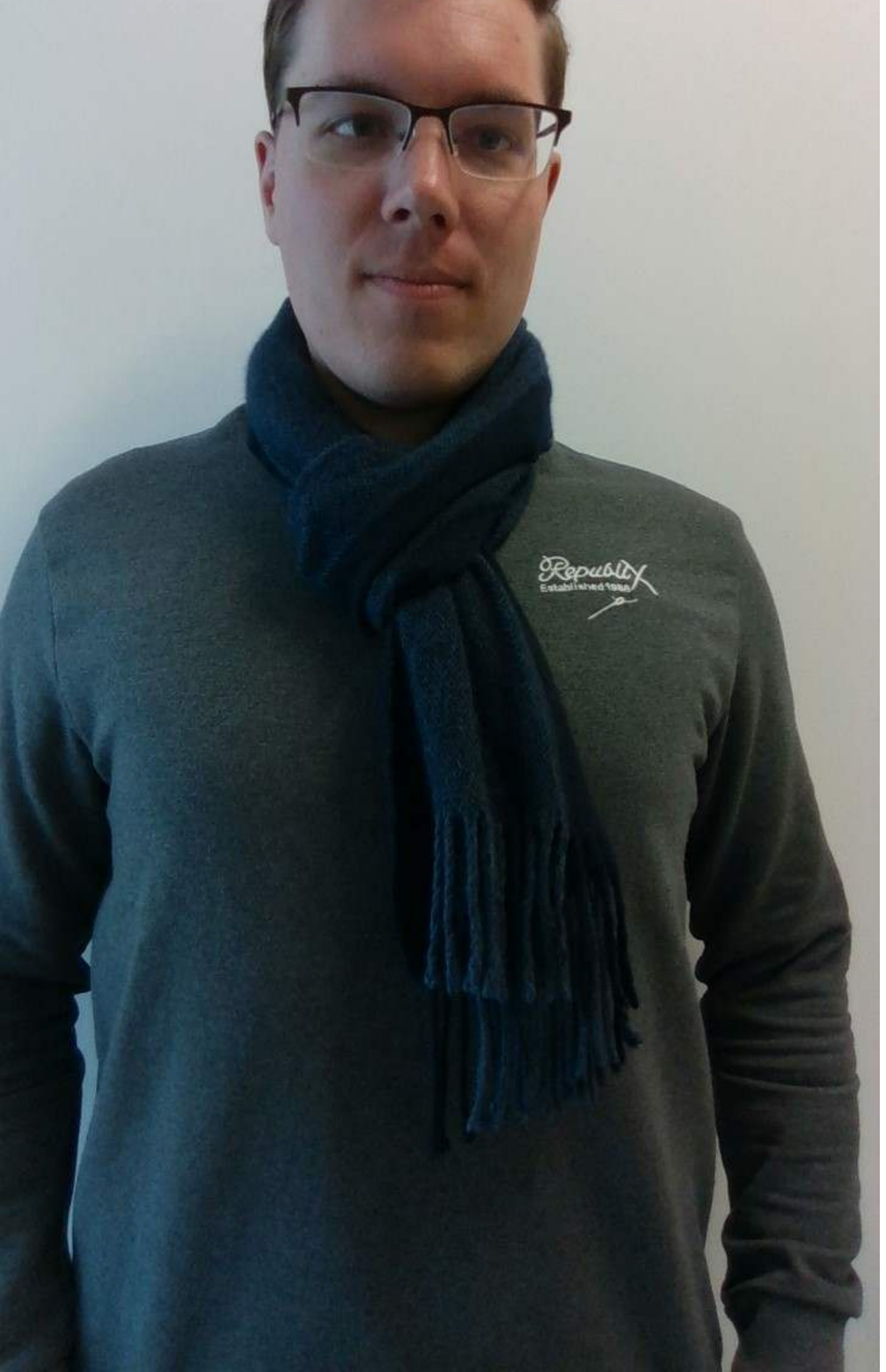}
  \caption{stretch}
\end{subfigure}
\begin{subfigure}{0.24\linewidth}
    \centering 
  \includegraphics[width=\textwidth]{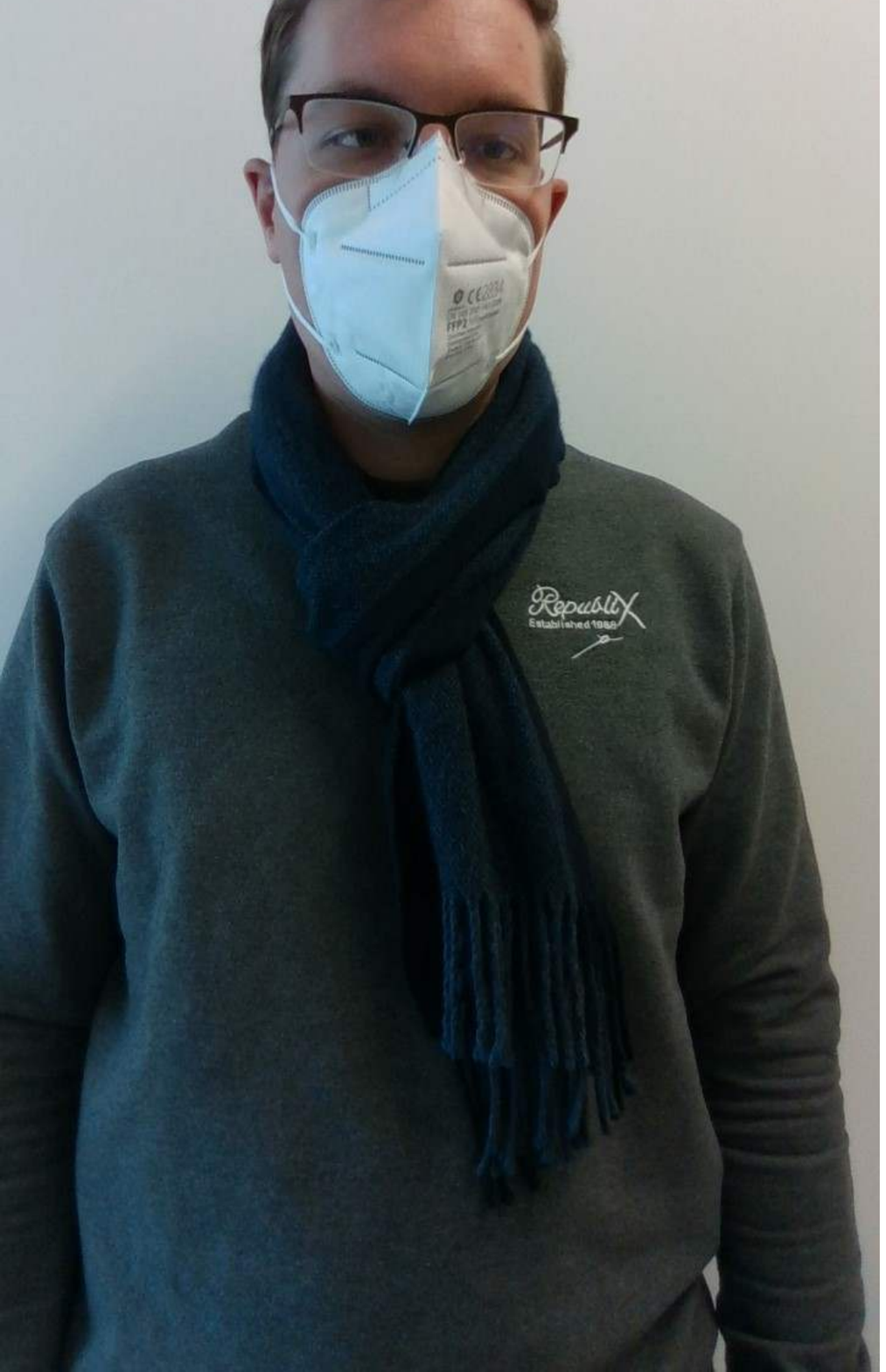}
  \caption{mask}
\end{subfigure}
\begin{subfigure}{0.24\linewidth}
    \centering 
  \includegraphics[width=\textwidth]{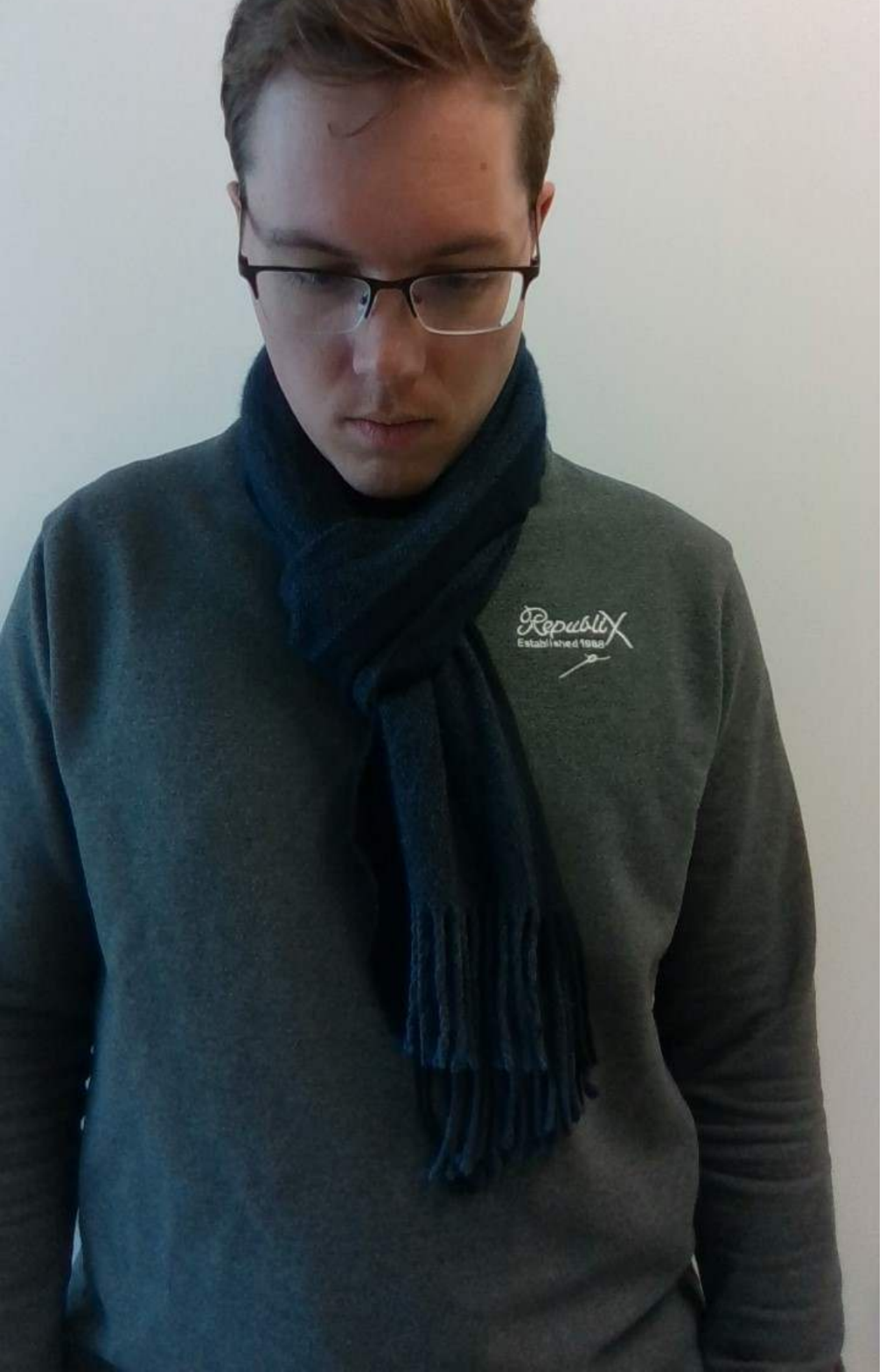}
  \caption{down}
\end{subfigure}
\begin{subfigure}{0.24\linewidth}
    \centering 
  \includegraphics[width=\textwidth]{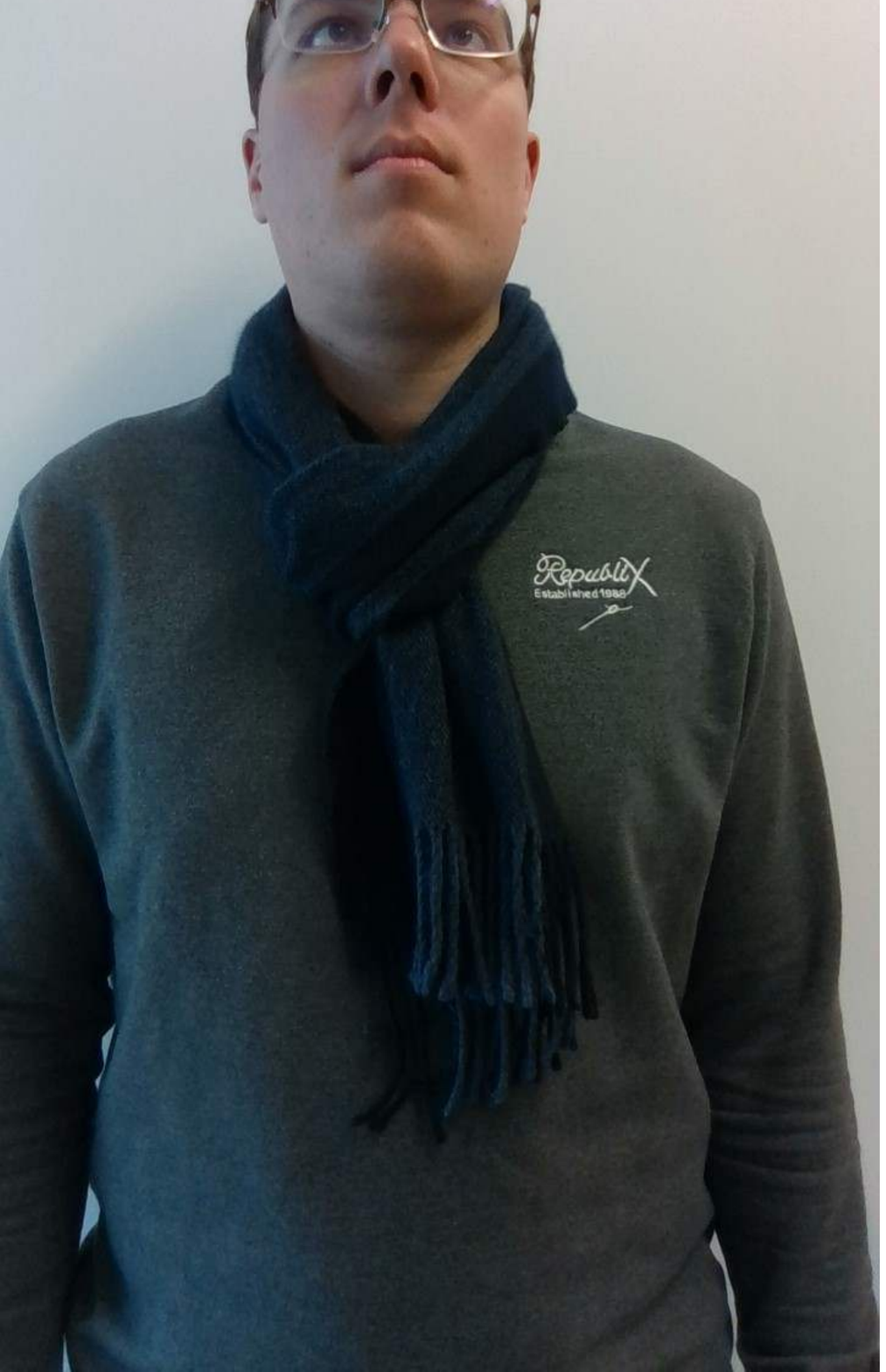}
  \caption{up}
\end{subfigure}
\caption{Example images of different bona fide presentations for a single data subject}
\label{fig:capturing_scenarios_bf}
\end{figure}

The data was captured in a bright room. A plain white wall served as the background and the D435 RGB-D sensor was mounted on a stable tripod. The subject was positioned at a distance so that the camera could shoot from the head to the waistline. Examples of bona fide presentations are depicted in Figure~\ref{fig:capturing_scenarios_bf}.

\section{Vulnerability Evaluation}

This section analyses the vulnerability of face recognition systems considering T-shirt PAs. Focus is mainly put on the feasibility of launching the attack which turns out to be more complex compared to other commonly used presentation attack instruments.

\subsection{Feasibility of the Attack}

\begin{table}[!htpb]
    \centering
\caption{Detection accuracy and average detection scores across algorithms and capturing scenarios for T-shirt and real faces~\cite{Ibsen-AttackingFaceRecognitionWithTshirts-IEEEAccess-2023}}
\label{tab:detection_performance_tshirt_attacks}
\begin{adjustbox}{max width=\textwidth}
   \begin{tabular}{@{\extracolsep{2pt}}llrrrrrrrr@{}} \toprule 
       & & \multicolumn{2}{c}{\textbf{dlib}}  &
      \multicolumn{2}{c}{\textbf{MTCNN}} & \multicolumn{2}{c}{\textbf{RetinaFace}}  &
      \multicolumn{1}{c}{\textbf{Avg.}} \\
      \cmidrule{3-4} \cmidrule{5-6} \cmidrule{7-8} \cmidrule{9-9}
Scenario & Face type  &  \multicolumn{1}{c}{Success \%} & \multicolumn{1}{c}{Avg. score}  & \multicolumn{1}{c}{Success. \%} & \multicolumn{1}{c}{Avg. score} & \multicolumn{1}{c}{Success \%} & \multicolumn{1}{c}{Avg. score} & \multicolumn{1}{c}{Success \%} \\
\midrule
   \multirow{2}{*}{Normal} 
   & real & 100 & 0.60 & 100 & 1.00 & 100 & 0.98 & 100  \\ 
   & T-shirt & 100 & 0.56 & 100 & 1.00 & 100 & 0.98 & 100  \\ 
   \midrule
      \multirow{2}{*}{Face covered} 
   & real & 0.50 & 0.01 & 10.45 & 0.41 & 11.94 & 0.67 & 7.63  \\ 
   & T-shirt & 98.01 & 0.50 & 99.50 & 1.00 & 100 & 0.97 & 99.17  \\    \midrule
      \multirow{2}{*}{Look left} 
   & real & 88.56 & 0.30 & 100 & 1.00 & 100 & 0.98 & 96.19  \\ 
   & T-shirt & 100 & 0.52 & 100 & 1.00 & 100 & 0.98 & 100  \\ 
   \midrule
     \multirow{2}{*}{Look right} 
   & real & 95.02 & 0.43 & 100 & 1.00 & 100 & 0.98 & 98.34  \\ 
   & T-shirt & 100 & 0.53 & 100 & 1.00 & 100 & 0.98 & 100  \\ 
   \midrule
         \multirow{2}{*}{Stretch T-shirt} 
   & real & 100 & 0.59 & 100 & 1.00 & 100 & 0.97 & 100  \\ 
   & T-shirt & 100 & 0.60 & 100 & 1.00 & 100 & 0.98 & 100  \\ 
   \midrule
      \multirow{2}{*}{Facial mask} 
   & real & 92.04 & 0.17 & 100 & 1.00 & 100 & 0.98 & 97.35  \\ 
   & T-shirt & 99.50 & 0.56 & 100 & 1.00 & 100 & 0.98 & 99.83  \\ 
   \midrule
      \multirow{2}{*}{Look down} 
   & real  & 88.56 & 0.34 & 100 & 1.00 & 100 & 0.98 & 96.19  \\ 
   & T-shirt & 100 & 0.56 & 100 & 1.00 & 100 & 0.98 & 100  \\ 
   \midrule
      \multirow{2}{*}{Look up} 
   & real & 89.55 & 0.31 & 99 & 0.99 & 100 & 0.99 & 96.18  \\ 
   & T-shirt & 100 & 0.55 & 100 & 1.00 & 100 & 0.98 & 100  \\ 
        \bottomrule
    \end{tabular}
\end{adjustbox}
\end{table}

To properly evaluate the impact of the T-shirt attacks on face recognition systems, we need to analyse how face detection algorithms deal with faces printed on T-shirts. If the faces on the T-shirts are not detected correctly or if the real face is detected with much higher confidence, the attack is likely to be filtered out at the pre-processing stage of the face recognition system and therefore cannot spoof the face recognition system. To evaluate this, three open-source face detection algorithms were used, namely RetinaFace~\cite{deng-retinaface-cvpr-2020}, MTCNN~\cite{Zhang-JointFaceDetectionAndAllignmentUsingMultitaskCascadedConvolutionalNetworks-2016} and dlib~\cite{King-MachineLearningToolkit-2009}. 

Looking at the width and height of each detected region of interest for the three different algorithms, we could see several outliers that indicate multiple faces that were inaccurately detected. Since we are only interested in regions with a face, these false detections were removed from the results.

To measure the detection performance, for each algorithm and a detected region, we record the detection confidence value of the algorithm, which is a measure of the probability that the region contains a face. We report the average confidence value for real and T-shirt faces for the three detection algorithms and eight detection scenarios. For each detection algorithm, a min-max normalisation of the confidence values is performed after filtering out the small faces. In addition, for each scenario and each recognition algorithm, we estimate how often the T-shirt face or the real face is correctly detected. Specifically, we refer to the success rate of a detection algorithm as the estimated proportion of correctly detected faces. We state this separately for real and T-shirt faces. The results for each scenario and detection algorithm are summarised in Table~\ref{tab:detection_performance_tshirt_attacks}. The results show that the T-shirt faces are successfully detected in almost all cases, with an average estimated detection rate for the three algorithms of $>99\%$ across all eight poses. The results also show that it is possible to prevent detection of the real face by occluding the face with the hands or, for some algorithms, increasing the chances of the attack by wearing a face mask or tilting the face. The results show that it is possible to carry out T-shirt attacks in semi-supervised or unsupervised scenarios. However, to increase the success rate, some efforts should be made to conceal the real face. 

\subsection{Attack Potential}

Once a face on a T-shirt is detected while the real face is concealed, it is likely that a match is returned by the face recognition system. This means that with high chance the similarity between the face printed on the T-shirt and the reference image against which it is compared to is above the systems decision threshold. This has already been showcased in~\cite{Ibsen-AttackingFaceRecognitionWithTshirts-IEEEAccess-2023}. It is argued that an attacker may even know the reference image, for instance in case he is in possession of an identity document used for comparison. In such a realistic scenario and using two face recognition systems, the IAPAR was reported to be about 99\% at the selected operating points of the systems.

\section{Attack Detection with Spatial Consistency Checks}

This section describes the proposed PAD method in detail and presents corresponding experiments on the previously described dataset. 

\subsection{Proposed Method}

Traditionally, face PAD methods solely analyse the facial region (cropped face image) in order to determine whether or not a PA has been launched against the face recognition system. PAD algorithms are usually designed to detect certain artefacts associated with specific presentation attack instruments, for instance moiré patterns on screen-based PAs. The usefulness of this type of PAD algorithms for detecting T-shirt PAs of the TFPA database has been investigated in~\cite{Ibsen-AttackingFaceRecognitionWithTshirts-IEEEAccess-2023}. Specifically, five PAD methods have been benchmarked on the TFPA database: the cross-modal focal loss algorithm proposed in~\cite{George-CMFLForRGBDFaceAntiSpoofing-CVPR-2021}, the vision transformed-based PAD method from~\cite{George-VIT-IJCB-2021}, a PAD method relying on the statistical differences between depth maps of bona fide images and T-shirt attacks, an anomaly detection approach trained on features only extracted from bona fide RGB images, and a fusion approach. The latter method achieved the best detection performance resulting in a detection equal error rate (D-EER) of approximately 12.5\%. While such a detection error rate appears to be high at first glance, the results are comparable to current state-of-the-art face PAD performance if the tested presentation attack instrument species are considered as unknown.

\begin{figure}[!tb]
\centering
\begin{subfigure}{0.24\linewidth}
    \centering 
  \includegraphics[width=\textwidth]{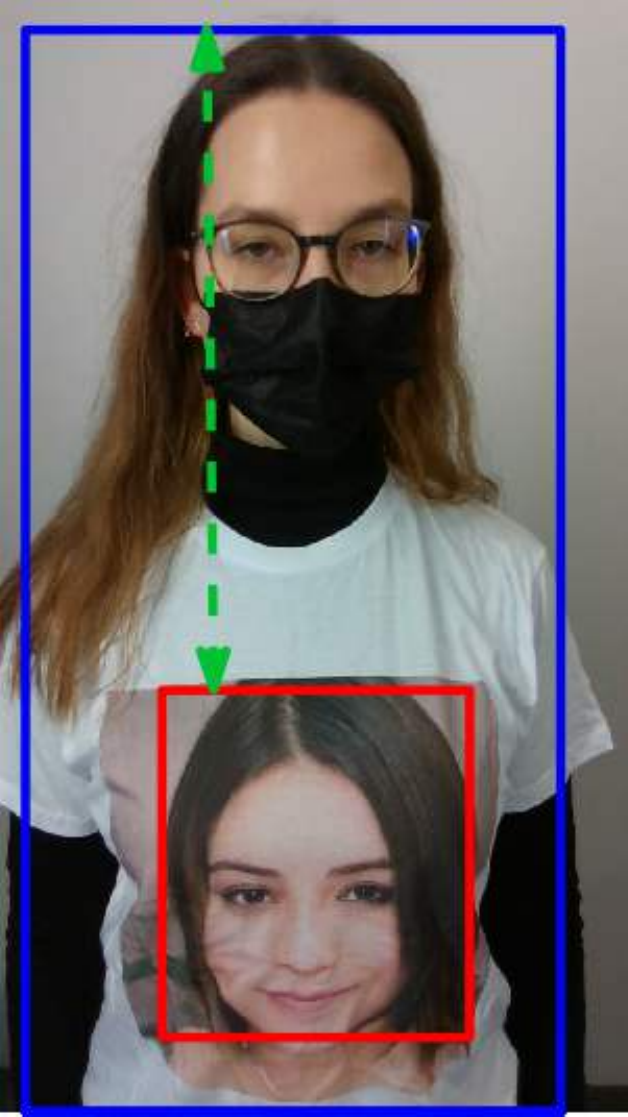}
  \caption{concealed real face}
\end{subfigure}\hspace{1cm}
\begin{subfigure}{0.24\linewidth}
    \centering 
  \includegraphics[width=\textwidth]{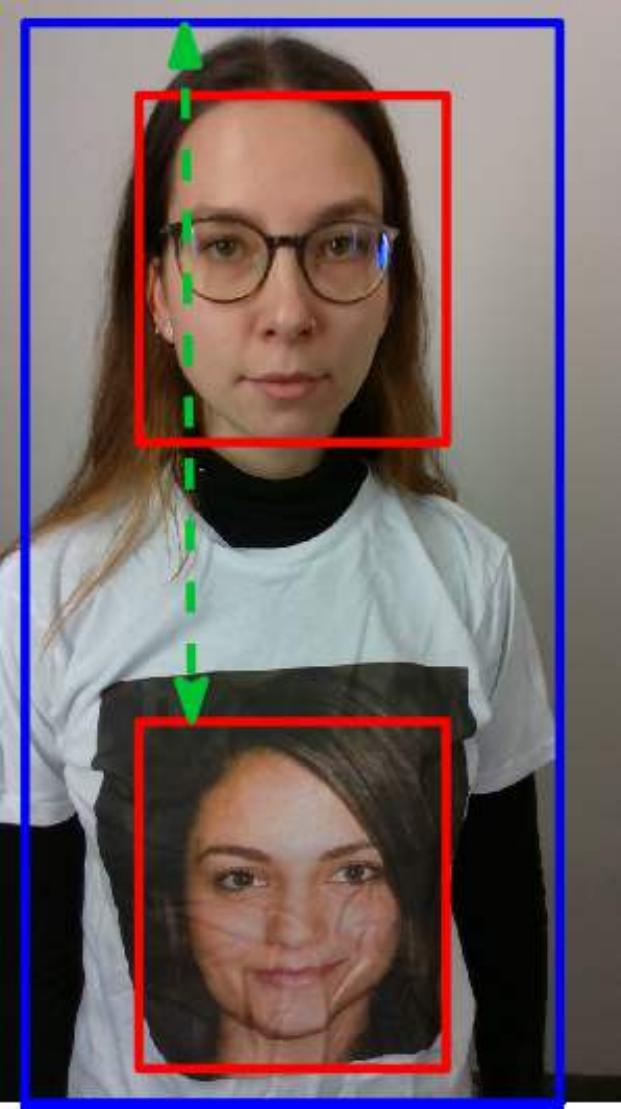}
  \caption{real face detected}
\end{subfigure}
\caption{Two examples in which the detected persons (blue bounding boxes) with the highest spatial position (according to its top left y-coordinate) are compared with the detected faces (red bounding boxes) with the lowest spatial position (according to its top left y-coordinate)}
\label{fig:examples_spatial_tshirt_detection}
\end{figure}

Instead of only analysing cropped face images, the idea of the PAD method proposed in this work is to employ the entire captured face image as input. Thereby, it is possible to analyse the position of a detected face with respect to the body of the detected person. To this end, face and person detection algorithms are applied, which return bounding boxes of detected faces and persons, respectively. It is assumed that only a single person is present in the captured face image sample. This assumption is realistic in numerous applications of face recognition; for instance, in automated border control where individuals present themselves to the capture device in a so-called man-trap such that no other person is visible in the background. Therefore, a T-shirt PA is detected if the difference of $y$-coordinates of the upper corners of the face bounding box with the lowest spatial position in the image (according to its top left y-coordinate) and the person bounding with the highest spatial position (according to its top left y-coordinate) differ beyond a predefined decision threshold. This difference is expected to be large if a T-shirt attack has been launched and the person detector is still able to detect the real person. Therefore, this methods work independent of whether the attacker is able to conceal his face or not, as illustrated in Figure~\ref{fig:examples_spatial_tshirt_detection}. A pseudo code of the proposed method is sketched in Algorithm~\ref{alg:spatial_consistency_check}, note that the code assumes that the origin of the image is in the top left corner.

\begin{algorithm}[!htb]
\caption{Generalised approach for spatial consistency checks}
\label{alg:spatial_consistency_check}
\begin{algorithmic}[1]
    \PROCEDURE[Origin of image is top left corner]{get\_spacial\_consistency\_score}{img} 
    \STATE person\_detections $\gets$ \textsc{find\_persons}(img)
    \STATE face\_detections $\gets$ \textsc{find\_faces}(img)

    \IF{\textsc{length}(face\_detections) $= 0$ or \textsc{length}(person\_detections) $= 0$} 
         \RETURN None \COMMENT{No face or person detected}
    \ENDIF

    \STATE persons\_y\_coords $\gets$ \textsc{list()} 
    \FOR{person $\in$ person\_detections}
            \STATE persons\_y\_coords.\textsc{add}(person.top\_left.y)
    \ENDFOR

    \STATE faces\_y\_coords $\gets$ \textsc{list()} 
    \FOR{face $\in$ face\_detections}
         \STATE faces\_y\_coords.\textsc{add}(face.top\_left.y)
    \ENDFOR

    \STATE \textsc{sort}(faces\_y\_coords, ascending $=$ True)
    \STATE \textsc{sort}(persons\_y\_coords, ascending $=$ True)

    \STATE lowest\_face\_y $\gets$ faces\_y\_coords[-1] \COMMENT{y-coordinate of the face furthest from the origin}
    \STATE top\_person\_y $\gets$ persons\_y\_coords[0]  \COMMENT{y-coordinate of the person closest to the origin}

    \RETURN $\frac{\text{lowest\_face\_y}}{\text{img.height}} - \frac{\text{top\_person\_y}}{\text{img.height}}$ \COMMENT{Normalised difference score}
    \ENDPROCEDURE
\end{algorithmic}
\end{algorithm}

The proposed algorithm can be combined with further rules. For instance, the detection of more than one face or person could already indicate an attack. Furthermore, the proposed method could be combined with traditional PAD methods that process cropped face images. This could be useful in the case where an attacker would manage to present a face printed on a T-shirt in front of his own face (similar to a paper-based PA).

\subsection{Detection Performance Results}
\begin{figure}[!h]
    \centering
    \includegraphics[width=\linewidth]{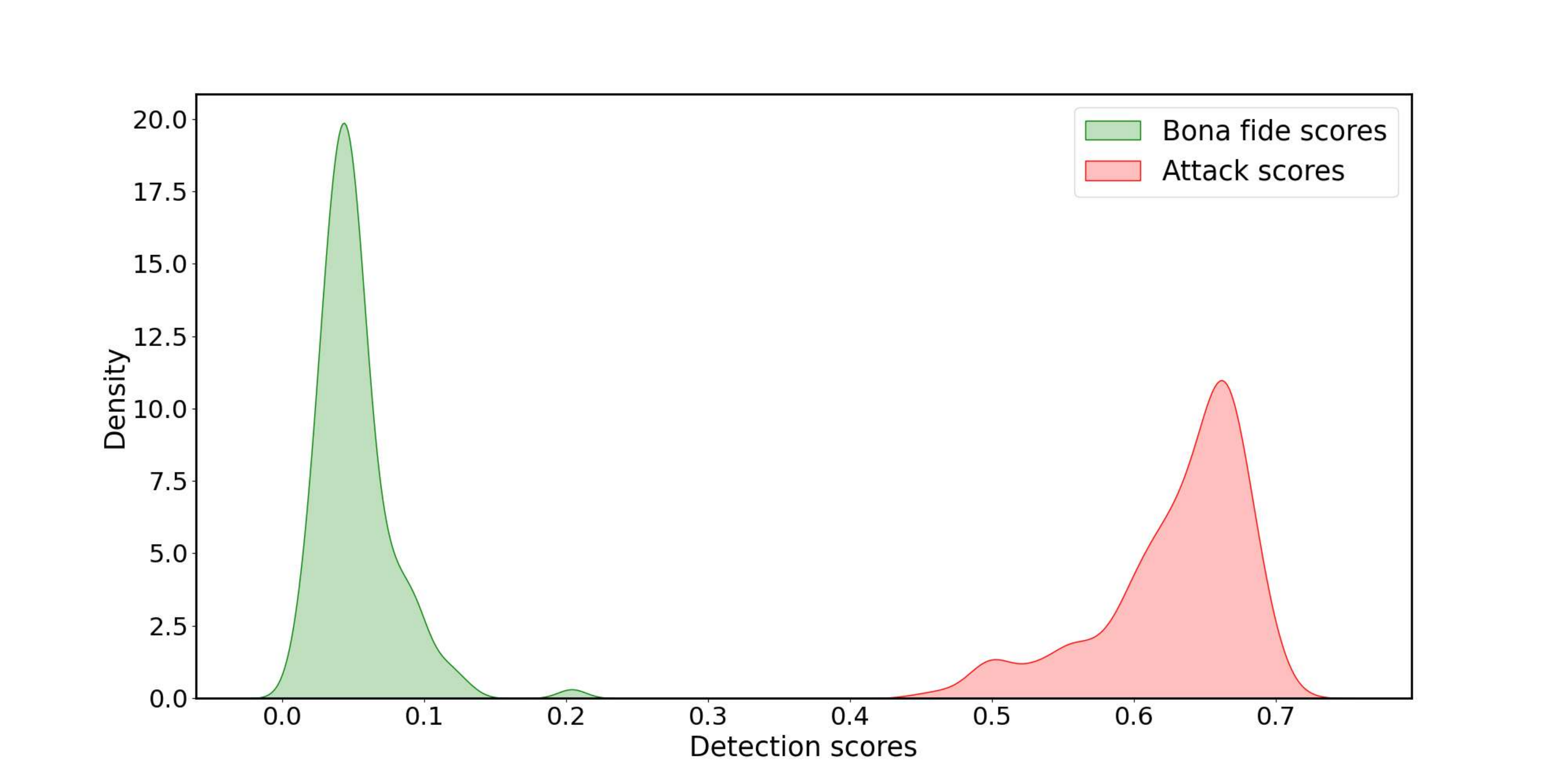}
    \caption{Detection scores obtained using the proposed spacial consistency checks on bona fide images and T-shirt PA images}
    \label{fig:scoredist}
\end{figure}

For the implementation of the PAD method we employed the RetinaFace algorithm for face detection since this algorithm revealed the best detection performance in the previous attack feasibility analysis. In addition, we use the YOLOv7 algorithm~\cite{wang2023yolov7} for person detection. Obtained score distributions for bona fide presentations and PAs are plotted in Figure~\ref{fig:scoredist}. A clear separation between the two classes is observed with optimal thresholds being around 0.3 -- 0.4. This means that the proposed detection algorithm based on spatial consistency checks works perfectly on the tested database. 

In contrast to previous PAD detection performance reported in~\cite{Ibsen-AttackingFaceRecognitionWithTshirts-IEEEAccess-2023}, where traditional algorithms were applied to the cropped face images, the presented algorithm is a viable mechanism to prevent against T-shirt PAs. The simplicity and effectiveness of the presented algorithm underline that face PAD algorithms can greatly benefit from analysing entire face image samples instead of pre-processed, i.e.\ segmented, image parts. Moreover, the proposed spatial consistency check can easily be integrated into the workflow of a face recognition system and combined with existing face PAD methods.  

\section{Conclusion}
With advances in deep learning and the availability of large facial databases, face recognition is becoming increasingly common. Nevertheless, it is clear that such systems are vulnerable to presentation attacks and must be accompanied by robust methods for detecting such attacks, which can otherwise be used to circumvent the security of a face recognition system. Recently, presentation attack detection methods have evolved towards generalisable presentation attack detection methods that are capable of detecting many different types of attacks, even attacks that were not seen during training. To accurately measure the capabilities of such presentation attack detection methods and advance the field, benchmarks that incorporate new and novel attack types need to be introduced. This work focuses on T-shirt impersonation attacks as a novel attack method that has received little attention in the scientific literature. The vulnerability of face recognition systems to these attacks was evaluated by analysing the feasibility of the attacks. The results showed that it was possible to launch the attack when the real face was obscured. In such cases, where the face on the T-shirt is recognised, the success rate of the attack is likely to be alarmingly high. To detect these attacks, we extended the TFPA database with 152 bona fide presentations and proposed a new detection method where state-of-the-art face and person detectors are combined to analyse the spatial positions of detected faces and persons based on which T-shirt attacks can be reliably detected. The results show perfect detection performance on the used database. Furthermore, the proposed algorithm can easily be combined with traditional presentation attack algorithms which work on cropped face images by simply applying these algorithms in cases where the spatial consistency check succeeds, i.e.\ no T-shirt attack was found by the proposed algorithm.  

\begin{acknowledgement}
This project has received funding from the European Union’s Horizon 2020 research and innovation programme under the Marie Sk\l{}odowska-Curie grant agreement No [860813]. Additionally, this work has been partially funded by the German Federal Ministry of Education and Research and the Hessian Ministry of Higher Education, Research, Science and the Arts within their joint support of the National Research Center for Applied Cybersecurity, ATHENE.
\end{acknowledgement}

\bibliographystyle{plain}
\bibliography{references}
\end{document}